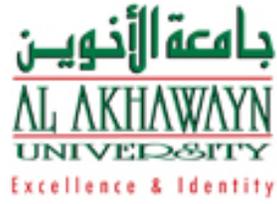

**Al Akhawayn University in Ifrane**

**School of Science and Engineering**

**Design and Engineering of a Robot Arm**

*by Ali Elouafiq*

*Supervised by Dr. Tajjedine Rachidi and Mr. Ahmed Elouafiq*

## I- ABSTRACT:

In the scope of the "Chess-Bot" project, this study's goal is to choose the right model for the robotic arm that the "the Chess-Bot" will use to move the pawn from a cell to another. In this paper, there is the definition and the structure of a robot arm. Also, the different engineering and kinematics fundamentals of the robot and its components will be detailed. Furthermore, the different structures of robotic arms will be presented and compared based on different criteria. Finally, a model for "the Chess-Bot" arm will be synthesized based on accurate algorithms and equations.

## II- INTRODUCTION:

Archytas, a Greek mathematician was the first to create an artificial flying pigeon; at 400BC the first robot was created. The word Robot was coined in 1920, coming from the Czech word ROBOTA which means "Compulsory Labor"[5]. Thereby, the first computer controlled hand was developed in MIT in 1961. Hence, robot arms, also known as robotic arms, become widely used in industry, and become very popular in research and development [5]. In the scope of the R&D project, the "Chess-Bot", I'm making a comparative study of different controlling arms, in order to choose a model for my project. In this paper, I'll present first the general structure of a robot arm, secondly its engineering fundamentals and measurements criteria and methods, thirdly the different models of robot arms, and finally the model that I'll base the "Chess-Bot" Controlling arms on it.

## II- THE STRUCTURE OF A ROBT ARM:

The robot arm, the most mathematically complex robot, goal is to find and interact with an object in the space, named **end effector**. The robot arm is widely used in many fields, such as industry, medicine, and in science and technology [2]. Thereby, there is a general structure that rules the entire robot arm, and puts a general scheme for this technology.

# Structure of a Robot arm  [2]

Ali Elouafiq Design and Engineering of a Robot Arm

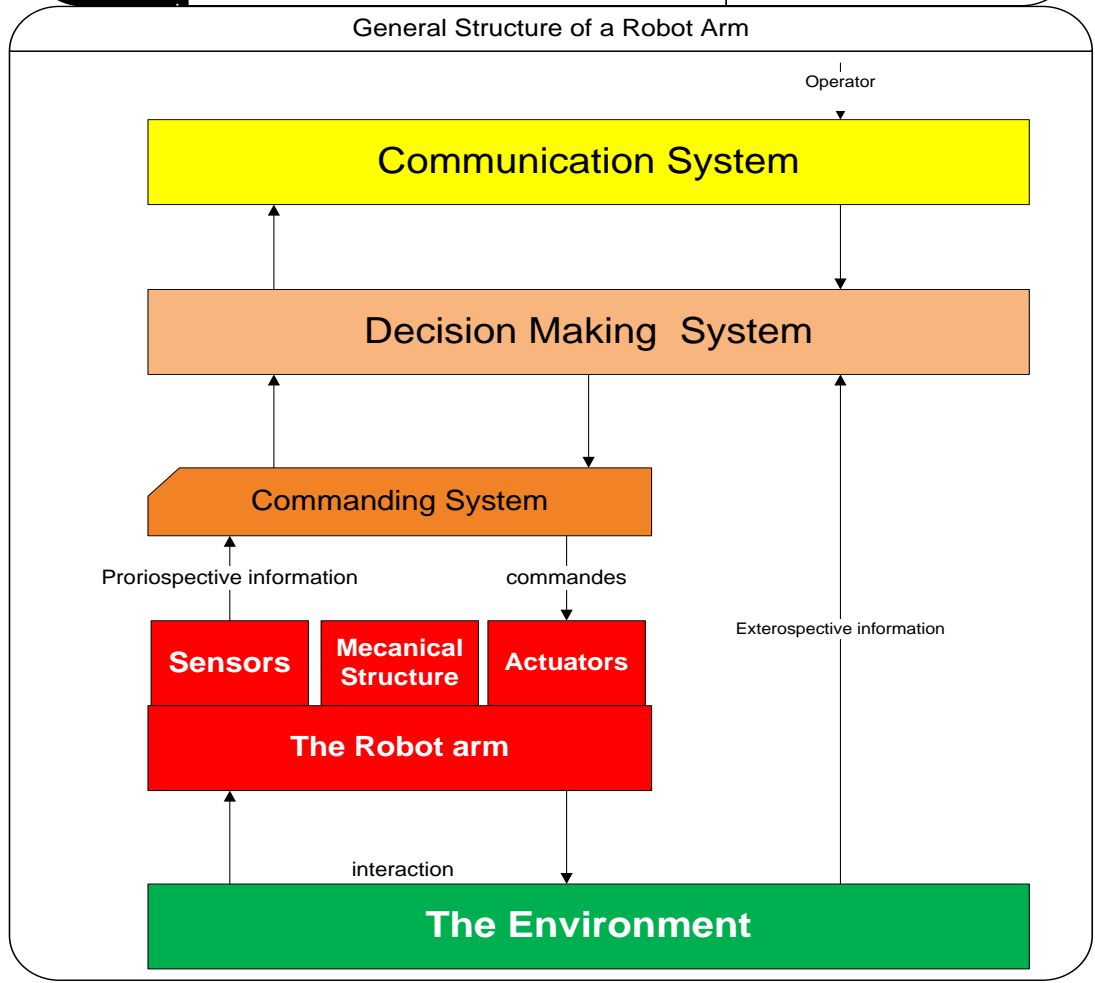

General Structure of a Robot Arm

Operator

**Communication System**

**Decision Making  System**

Commanding System

Proriospective information          commandes

Sensors    Mecanical Structure    Actuators

**The Robot arm**

Exterospective information

interaction

**The Environment**

Description:
   Mechanical Structure: the structure that supports the end effector
   Actuators: acts on the mechanical structure and modifies the tend effector.
   Sensors: necessary for the command and for the decision making. There are two kinds of sensors, **proprioceptive** and **exteroceptive**. The **proprioceptive** sensors enable to know the mechanical state of the robot arm. The **exteroceptive** gives the stat of the environement of the robot.
   Commanding System: drives the actuators of the manipulator based on the decision made by the « Decision Making System » and the proprioceptive sensors.
   Decision Making System:  is the CPU of the whole system that makes decision based on the input of the exteroceptive sensors and the commanding system, and give output to the communication system.
   Communication System: a system that communicates to the end operator about the what happens inside the system, via a G.U.I.

*(This scheme is taken and translated from [2])*

### III- ENGINEERING FUNDAMENTALS:

#### III-A- PARAMETERS AND MECHANISMS:

##### III-A-1 PARAMETERS:

- In order to define the status of a free object in space, we can fix 3 non-aligned points of that object; in other words, nine dependant parameters, since the coordinates of those points are linked by 3 relations that describe the distance between them.[2]
- Thereby, in order to define the situation of a free object, we need to know six (nine minus three) independent parameters [2] :
    - Three independent parameters that defines the position of a point from that object. (Such as Cartesian coordinates or cylindrical coordinates).
    - Three other independent parameters that determines the orientation of the object according to the previous point. (Such as Euler angles or Euler parameters)

##### III-A-2 JOINS AND LINKAGES:

- A robot mechanism is a multi-body system with the multiple bodies connected together. We begin by treating each body as rigid, ignoring elasticity and any deformations caused by large load conditions. Each rigid body involved in a robot mechanism is called a link, and a combination of links is referred to as a linkage. In describing a linkage it is fundamental to represent how a pair of links is connected to each other. There are two types of primitive connections between a pair of links, as shown in Figure III-A-2-1. The first is a prismatic joint (P) where the pair of links makes a translational displacement along a fixed axis. In other words, one link slides on the other along a straight line. Therefore, it is also called a sliding joint. The second type of primitive joint is a revolute joint (R) where a pair of links rotates about a fixed axis. This type of joint is often referred to as a hinge, articulated, or rotational joint.[1,2]

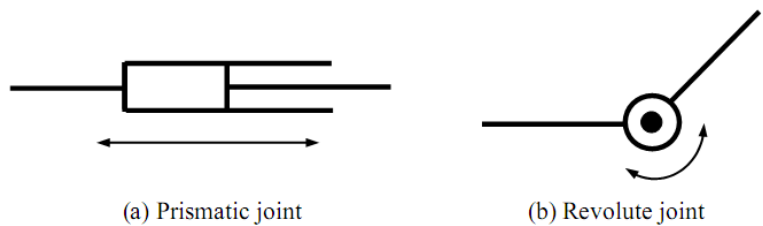

(a) Prismatic joint                    (b) Revolute joint

*Figure III-A-2-1: Primitive joint types: (a) a prismatic joint and (b) a revolute joint*
*(Figure credited to [1])*

- Thus, in order to create useful robotic mechanisms for manipulation and locomotion. These two types of primitive joints are simple to build and are well grounded in engineering design. Most of the robots that have been built are combinations of only these two types.[1]

### III-B- Degree of Freedom:

#### III-B-1 Definition:

- The *degree of freedom, or DOF,* is the number of independent parameters that fixes the situation of an end effector. It is a very important term to understand. Each degree of freedom is a joint on the arm, a place where it can bend or rotate or translate. You can typically identify the number of degrees of freedom by the number of actuators on the robot arm. When building a robot arm we should have as few degrees of freedom allowed for the application. Since, each degree requires motor, an often (not all the time), and exponentially complicated algorithms and cost.[2,7]

#### III-B-2 The Denavit-Hartenberg Convention:

- The Denavit-Hartenberg (DH) Convention is the accepted method of drawing robot arms in free body diagrams, where a join can only do two kind of motions: translate or rotate, in a the three-dimensional space (*o.* x, y ,z). The end effectors' DOF (also known as the gripper) is not mandatory to show, because it is often complex with multiple DOF, so for simplicity it is treated as separate in basic robot arm design. [7]

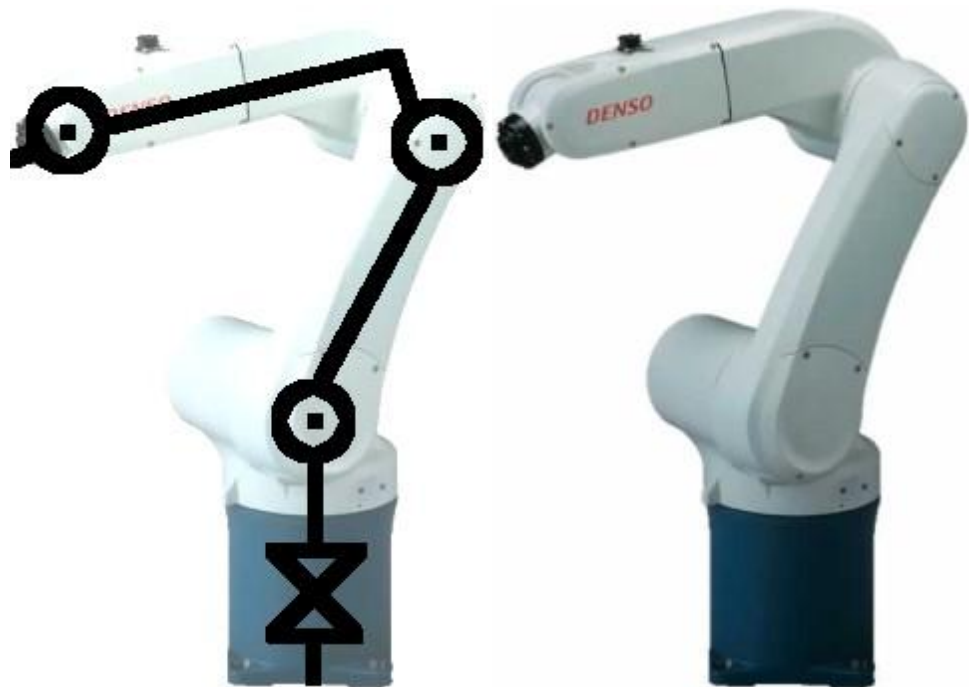

*Figure III-B-2-a: 4 DOF Robot Arm*
*(Figure credited to [7])*

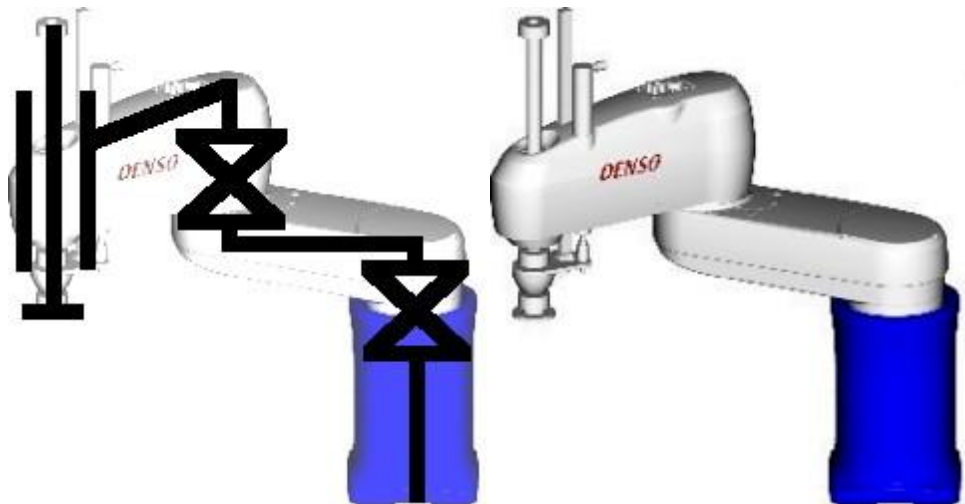

*Figure III-B-2-b: 3 DOF Robot Arm, with a translation joint*
*(Figure credited to [7])*

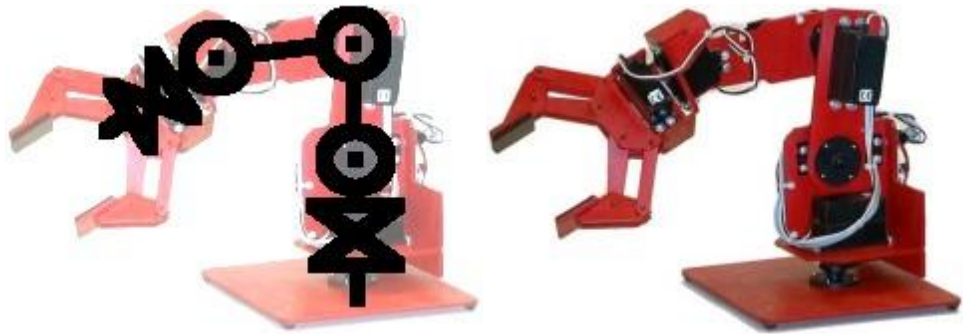

*Figure III-B-2-c: 5 DOF Robot Arm*
*(Figure credited to [7])*

- It is important to note that DOF has its limitations, known as the **configuration space**. Since not all joints can revolve 360 degrees, each join has a max angle restriction. Limitations could be from wire wrapping, actuator capabilities, servo max angle or other design restrictions. It is very important to label each link length and joint max angle on the free body diagram.[7]

### THE DEGREE OF FREEDOM OF A TASK:

- The degree of freedom of a task is the number of independent parameters that allows fixing a specific situation for the end effector. Its symbol is DOF$t$. In order to achieve a task, it is necessary that DOF$t$= DOF$r$ *(Degree of Freedom of the robot).* This is the first necessary condition to find the end effector; also there should be a possible configuration to find the end effector.[2,7]

### III-C- ROBOT WORKSPACE:

- The **reachable space** or the robot workspace is the physical space where gripper can move and have a position, in other words, the space composed by all the points that the gripper can reach. The workspace is dependent on[2,7]:
    - ○ The DOF angle/translation limitations.
    - ○ The arm link lengths.
    - ○ The angle at which something must be picked up at.
    - ○ The main dimensions of the Robot, which is the maximum lengths of the parts of the robot.
    - ○ The robot configuration.
    - **Example:**
- For the reason that there are too many robotic arms configuration, we will choose a simple case (the one shown bellow to explain the other principles).

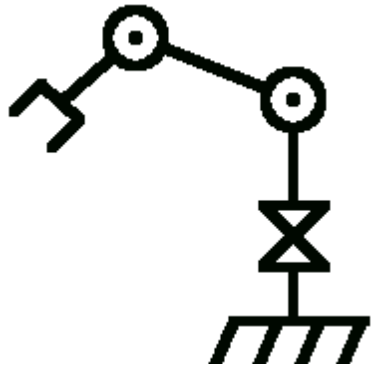

*Figure III-C-1: 3DOF simple robotic arm.*

*(Figure credited to [7])*

- We assume that all joints rotate a maximum at an angle $\theta = \pi$ (for the reason that all servo motors cannot exceed that limit). In order to find the workspace we should trace all locations that the end effector can reach as in the image below.

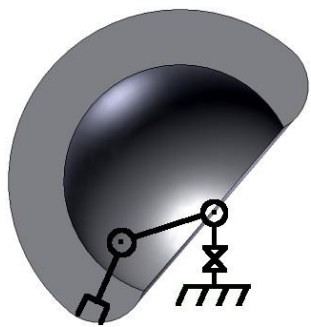

*Figure III-C-2 : Workspace of a simple 3DOF robotic arm.*

*(Figure credited to [7])*

- There are other wide used robot arm models, these are example of their workspace:

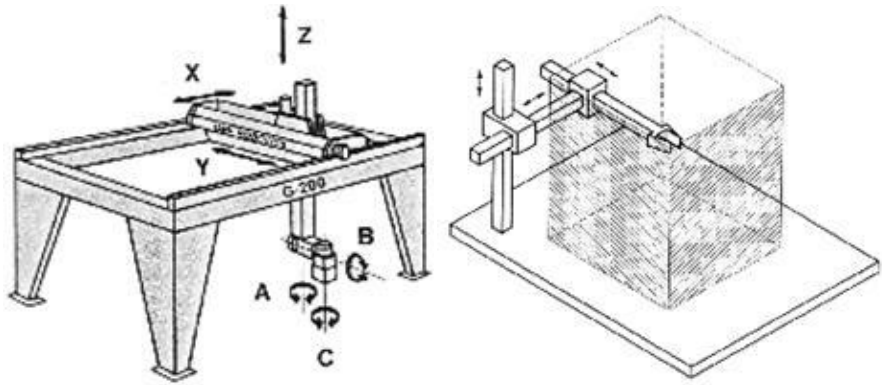
*Figure III-C-3: Cartesian Gantry Robot Arm*

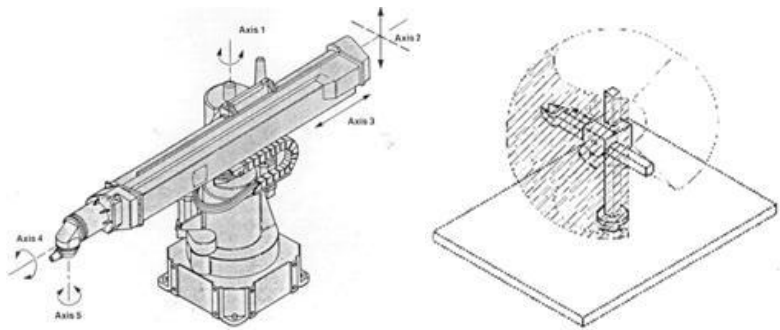
*Figure III-C-4: Spherical Robot Arm*

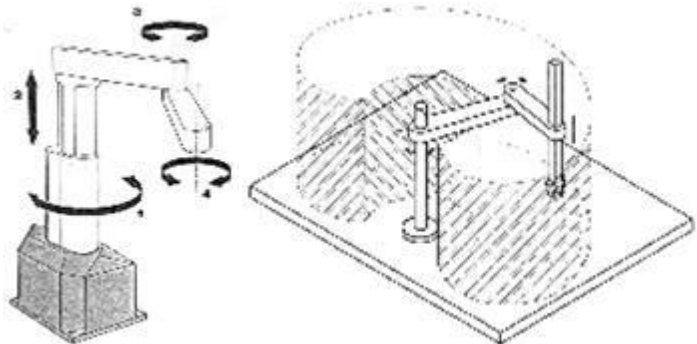
*Figure III-C-5: SCARA Robot Arm*

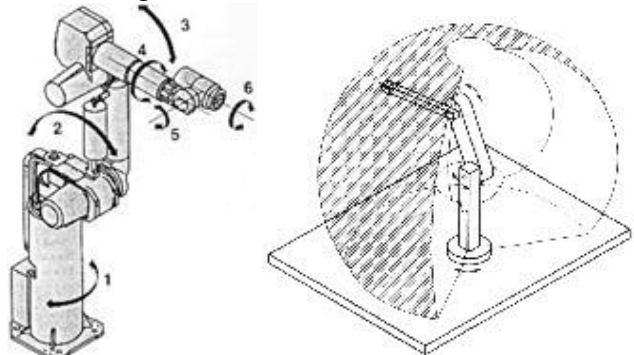
*Figure III-C-6: Articulated Robot Arm*
*(Figures credited to [7])*

### III-D- BASE STRUCTURE MODELS

- These are different base structure models that are widely used in robot arm modeling; each one has some specific usage and specific design. The (P) stands for the prismatic join and the (R) for the revolute join.

### III-D-1 PPP STRUCTURE:

- This model is used by 14% of the Industrial robot arms, such as RENAULT "P80" and HI-T-HAND2, and it is well adapted to the Cartesian coordinate configuration[2].

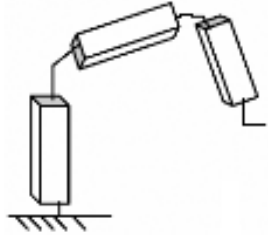

*Figure III-D-1-a PPP Structure*
*(Figure Credits to Ali Elouafiq)*

### III-D-2 RPP OR PRP STRUCTURE:

- This model concerns 47% of the industrial robot arms, such as VERSATRAN "E series", and it is very well adapted to the cylindrical coordinate configuration[2].

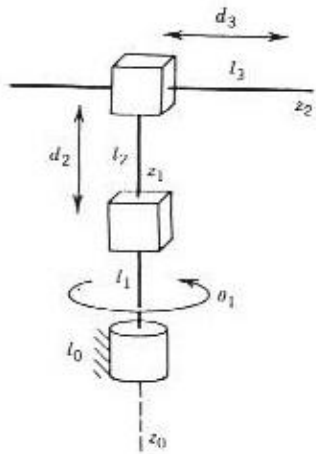

*Figure III-D-2-a PPR Structure*
*(Figure credits to [3])*

### III-D-3 RRP Structure:

- This model is present by 13% of the industrial robot arms, such as POLAR 6000 and TOSMAN, and it is well adapted to spherical coordinate configuration.[2]

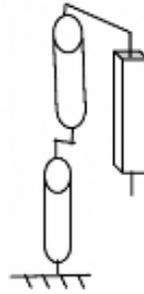

*Figure III-D-3-a RPR or PRR Structure*

*(Figure Credits to Ali Elouafiq)*

### III-D-4 RPR or PRR Structure:

- This type of structure is only used by 1% of the industrial robot arms, such as REIS 625, and it is well adapted to torical coordinate configuration.[2]

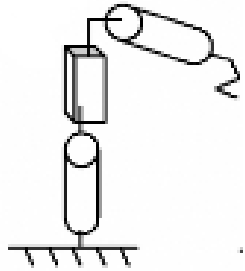

*Figure III-D-3-a RPR or PRR Structure*

*(Figure Credits to Ali Elouafiq)*

### III-D-5 RRR Structure:

- This model is used by 25% of the industrial robot arms, such as ASEA IRb6 and SCEMI 6P01, since it is well adapted to anthropomorphic coordinates that are similar to human arms.[2]

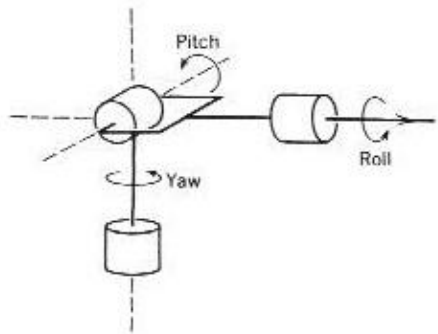

*Figure III-D-5-a RRR Structure*

*(Figure credits to [3])*

- In order to select the convenient motor, we should do first the force calculation to make a well designed decision. One should make sure to not forget, in the calculations, the maximum weight of the end effector and the weight object carried.  As a first step a free body diagram should be designed with the robot arm stretched out to its maximum length. [7]

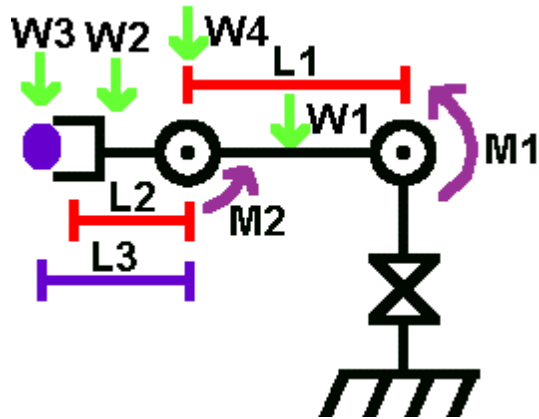

*Figure III-E-1: Free body diagram of a 3DOF Arm*
*(Figure credits to [7])*

- The parameters that should be taken into consideration are :
  o weight of each linkage
  o weight of each joint
  o weight of object to lift
  o length of each linkage.

- A moment arm calculation should be done by multiplying downward force times the linkage lengths. This calculation must be done for each lifting actuator. This particular design has just two DOF that requires lifting, and the center of mass of each linkage is assumed to be Length/2.
- In the figure III-E-1 we have:

  Torque about Joint 1: $M1 = \frac{L1}{2}.W1 + L1.W4 + \left(L1 + \frac{L2}{2}\right).W2 + (L1 + L3).W3$

  Torque about Joint 2: $M2 = \frac{L2}{2}.W2 + L3.W3$

- We see that for each DOF we should add, the equations gets more complicated, and the joint weights get heavier. The reason why we should have an optimized design at the beginning.[7]

### III-F- Kinematics:

#### III-F-1 Forward Kinematics:

- The forward kinematics is the method for determining the orientation and position of the end -effector, given the joint angles and link lengths of the robot arm. The example shown bellow is the one of the 3DOF robot shown in the Figure III-C-1. The Effector location can be calculated with given joint angles and link. [7]

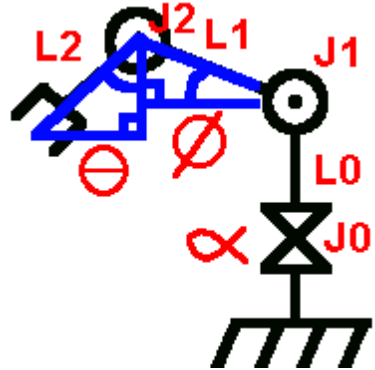

*Figure III-F-1-a shows the different representation on the 3DOF robot (Figure credits to [7])*

- We Assume that the base is located at x=0 and y=0. The first step would be to locate x and y of each joint. [7]
- Joint 0 (with x and y at base equaling 0):
$$x_0 = 0 \ and \ y_0 = L_0$$
- Joint 1 (with x and y at J1 equaling 0):

$$cos(\Phi) = \frac{x_1}{L_1} \rightarrow x_1 = L_1. cos(\Phi)$$
$$sin(\Phi) = \frac{y_1}{L_1} \rightarrow y_1 = L_1. sin(\Phi)$$

- Joint 2 (with x and y at J2 equaling 0):
$$cos(\Theta) = \frac{x_2}{L_2} \rightarrow x_2 = L_2. cos(\Theta)$$
$$sin(\Theta) = \frac{y_2}{L_2} \rightarrow y_2 = L_2. sin(\Theta)$$

- End Effector Location:

$$Xf = x_0 + x_1 + x_2 = L_1. cos(\Phi) + L_2. cos(\Theta)$$
$$Yf = y_0 + y_1 + y_2 = L_0 + L_1. sin(\Phi) + L_2. cos(\Theta)$$
$$z = \alpha \ (in \ cylindrical \ coordinates)$$

- The angle of the end effector $\Omega = \Phi + \Theta$ [7]

### III-F-2 Inverse Kinematics: [7]

- Inverse kinematics is the opposite of forward kinematics. This is used when we have a desired end effector position (Xf, Yf), but need to know the joint angles required to achieve it. This is the most used kinematic calculation; unfortunately, this calculation is much more complicated. As such, we will just focus on this specific design example in the figure III-C-1.
- We have:

$$\Phi = cos^{-1}\ (\frac{x^2 + y^2 - L1^2 - L2^2}{2 \cdot L1 \cdot L2})$$

$$\Theta = sin^{-1}\ (\frac{y \cdot (L1 + L2 \cdot c2) - x \cdot L2 \cdot s2}{x^2 + y^2})$$

Where:

$$c2\ = \frac{(x^2 + y^2 - L1^2 - L2^2)}{2 \cdot L1 \cdot L2}\ \textbf{and}\ s2\ = \sqrt{1\ -\ c2^2}$$

- The problem of inverse kinematics is that it requires non-linear simultaneous equations and, for some case there is the possibility to have some inconsistencies in the solution such as multiple number of solutions, sometime, infinite number of solutions, which requires some adjustments in the algorithms to choose optimal solutions, based on torque or previous arm position.
- But when there are zero solutions, thus the location required is outside the workspace; however, sometimes it is inside the workspace but cannot be gripped at a possible angle.
- Other errors that occur are name **singularities;** it is a place for infinite acceleration, which can confuse equations or leave give errors to the motor, since they cannot achieve infinite accelerations.
- Finally, another design decision that should be taken into consideration, no need to have advanced equations on a tiny processor, since exponential equations should be simplified, since it will take a forever to calculate them on a microcontroller.

### III-G- Velocity: [7]

- It is mathematically complex to calculate the end effector velocity. Therefore a step by step method should be followed to make the calculations are as simplified as possible. We first assume that the robot arm (held straight out) is a rotating wheel of L diameter. The joint rotates at Y rpm(radian per meter), so therefore the velocity is :

$$V = 2\,\pi.R.f$$

- Nevertheless, the end effector does not just rotate about the base, but can go in many directions. The end effector can follow a straight line, or curve, etc.
- With robot arms, the quickest way between two points is often not a straight line. If two joints have two different motors, or carry different loads, then max velocity can vary between them. When you tell the end effector to go from one point to the next, you have two decisions. Have it follow a straight line between both points, or tell all the joints to go as fast as possible - leaving the end effector to possibly swing wildly between those points.
- In the image below the end effector of the robot arm is moving from the blue point to the red point. In the top example, the end effector travels a straight line. This is the only possible motion this arm can perform to travel a straight line. In the bottom example, the arm is told to get to the red point as fast as possible. Given many different trajectories, the arm goes the method that allows the joints to rotate the fastest.

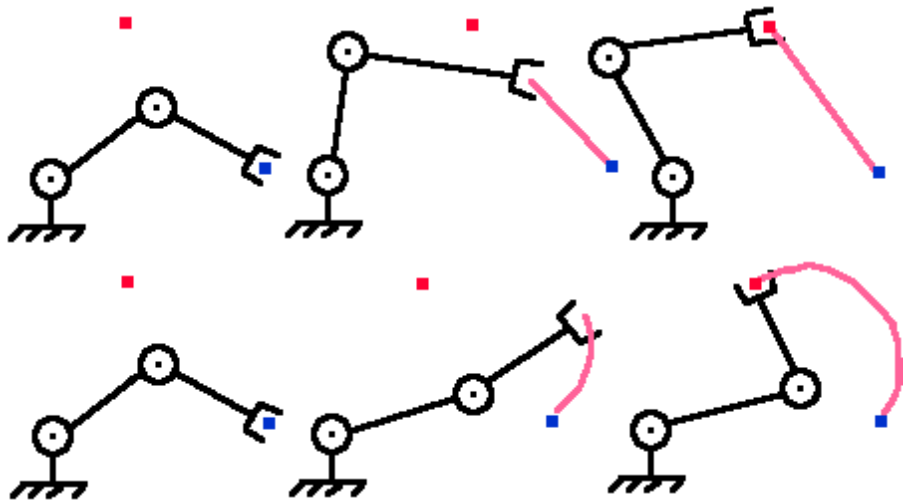

Figure III-G-1: Robot arm using translation or rotation to move between two points (Figure credits to [7])

- To decide on which method are better, many deciding factors interfere. Usually we use straight lines when the object the arm moves is heavy, since it requires the momentum change for movement (**momentum = mass * velocity**). But for maximum speed, when the arm is not carrying anything, or light weights, better to use joins rotations.

- When we want a robot arm to operate at a certain rotational velocity, we should calculate how much torque would a joint need. First, let's go back to our free body diagram in figure

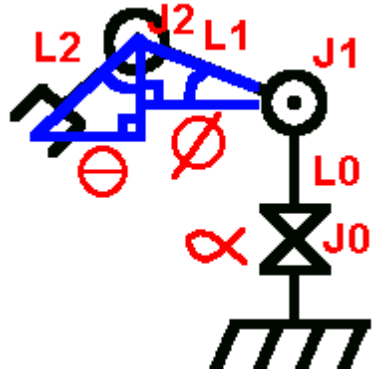

Figure III-F-1

- Suppose we want joint J0 to rotate π in 2 seconds. We know that J0 is not affected by gravity, so all we need to consider is momentum and inertia. Putting this in equation form we get this:

$$\tau = I * \omega = M.d^2.\frac{d\omega}{dt} = M.d^2.\frac{(\omega_1 - \omega_0)}{dt} = M.d^2.\frac{d\alpha}{dt^2}$$

$$when \ t = 0 \rightarrow \omega_0 = 0 \rightarrow \tau = M.d^2.\frac{\omega}{t}$$

$$\rightarrow \tau_{arm} = M_{arm}.\frac{L^2}{4}.\frac{\omega}{t} \ and \ \tau_{obj} = m.L^2.\frac{\omega}{t}$$

$$\tau_{motor} = \tau_{arm} + \tau_{obj} = (\frac{M_{arm}}{4} + m)L^2.\frac{\omega}{t}$$

### III-I- ACTUATORS [1]

- Actuators are one of the key components contained in a robotic system. A robot has many degrees of freedom, each of which is a servoed joint generating desired motion. We begin with basic actuator characteristics and drive amplifiers to understand behavior of servoed joints.

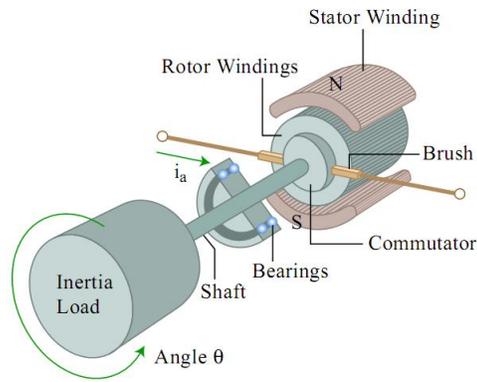

*Figure III-I-1: Construction of DC motor*

*(Figure credits to [1])*

- Figure III-I-2-a illustrates the construction of a DC servomotor, consisting of a stator, a rotor, and a commutation mechanism. The stator consists of permanent magnets, creating a magnetic field in the air gap between the rotor and the stator. The rotor has several windings arranged symmetrically around the motor shaft. An electric current applied to the motor is delivered to individual windings through the brush-commutation mechanism, as shown in the figure. As the rotor rotates the polarity of the current flowing to the individual windings is altered. This allows the rotor to rotate continually.[1]

- The DC motor described previously is the simplest, yet efficient motor among various actuators applied to robotic systems. Traditional DC motors, however, are limited in reliability and robustness due to wear of the brush and commutation mechanism. In industrial applications where a high level of reliability and robustness is required, DC motors have been replaced by brushless motors and other types of motors having no mechanical commutator.[1]

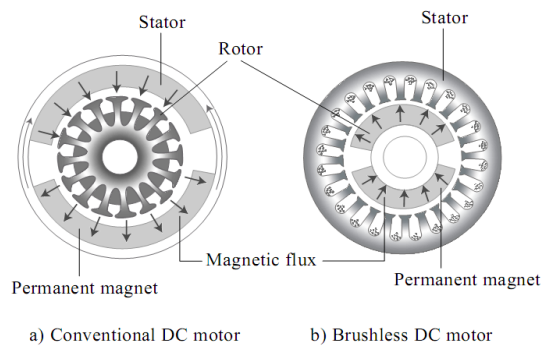

*Figure III-I-2: Construction of brushless DC motor and conventional DC motor*

*(Figure credits to [1])*

## V- CHOOSING A MODEL FOR THE CHESS-BOT:

After studying the needs of the project "the ChessBot", I've synthesized a design of the robot arm that will enable moving up chess pawns from a coordinate to another. I'll present in this section the full design of the arm that includes the structure, the kinematic formulations, the end effector design, and finally the algorithms that will drive the arm.

The arm robot has an engine, which receives queries and translates them into arm motion, and gripper actions. The queries should be in the following forms:

| query | Description |
| --- | --- |
| go to **x** | Moves the arm to **x** position |
| go to **y** | Moves the arm to y position |
| move to [**x ,y**] | Moves the arm to **x** and **y** position |
| move from [**x ,y**] to [**x1,y1**] | Moves an object from **x, y** to **x1,y1** |
| return to o | Returns to its initial position |
| grab | Grabs the object with the gripper |
| release | Releases the gripper |

### V-1 THE STRUCTURE:

The robot arm that will be moving pawns on the chess board will have 4 Degrees of freedom. Thereby, the structure will be **RRRR**. The Robot arm calculations results in Cartesian coordinates, since we deal with chess board cells.

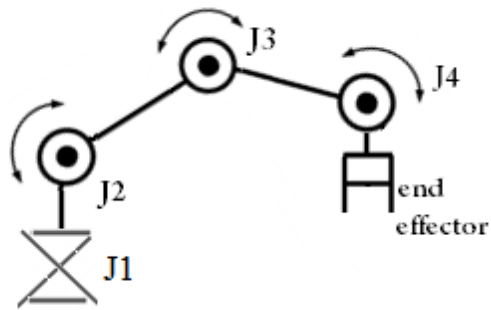

Figure V-1-a: the PRRR structure of the robot arm according to Denavit-Hartenberg Convention

(Figure credits to Ali Elouafiq)

For easier coordinate calculation, and rapid computation, the J1 join is revolute which enable us to move easily with x and y coordinates. The first revolute join (J2) is used to calibrate the z coordinates of the end effector. The second revolute join's (J3) role is to calibrate the y coordinates of the end effector. And the last join (J4) is adjusting the position of the end effector so, that it is always looking downward.

In order to find the right kinematics equations we should first model the environment where the robotic arm will interact. The environment is actually the chess board, the figure bellow show how the chessboard will be represented to the arm. The Side length of the board is **L, *dif*** is the distance between the chess board and the center of the robot.

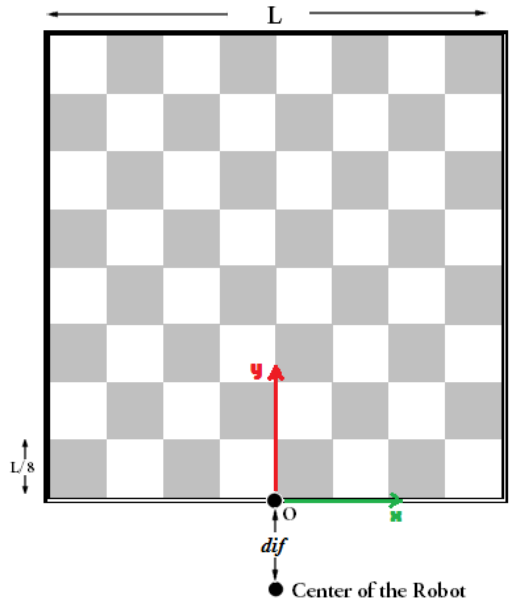

Figure V-2-a: the Chess Board where the events will occur.

(Figure credits to Ali Elouafiq)

The point **O** is the origin of the Cartesian coordinates, all the cells at its left have negative **x** parameters. Each, cell has a side length **L/8**, and a corresponding **(x,y)** coordinates that corresponds to its center**.**

**(Xf,Yf)** refers to the actual locations of x and y on the board and it requires their formulas are as follow:

$$\left(X_f, Y_f\right) = (x.\frac{L}{8} - \frac{L}{16}, y.\frac{L}{8} - \frac{L}{16} + dif)$$

The robot arm has 4 links, one that corresponds to each join:

$$\{1 \leq i < 5, L_i \text{ links } J_i \text{ to } J_{i+1}, \text{assuming that } J_5 \text{ is the gripper}\}$$

$L_0$ has length ***h***, $L_1$ and $L_2$ has the same length ***d***, $L_4$ has length ***h-$D_G$*** (the gripper length).

The $X_f$ is found easily by the translation of the prismatic join **J1**. But to calibrate $Y_f$, more calculations based on **J2** and **J3**. Let's consider the following triangle.

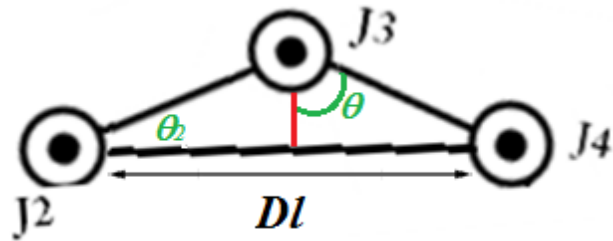

Figure V-2-b: the triangle formed by the 3 joins

(Figure credits to Ali Elouafiq)

In the figure θ is the angle by which **J3** rotates, θ2 is the angle by which **J2** rotates. The join **J1** rotates with an angle θ1. As a result, we will have the following inverse kinematics equations:

$$d = \frac{15L - 16dif}{32} \ \& \ Dl = \sqrt{Xf^2 + Yf^2}$$

$$\theta_1 = tan^{-1}\left(\frac{Xf}{Yf}\right) \ (the\ rotation\ related\ to\ J1)$$

$$\theta_2 = sin^{-1}\left(\frac{Yf}{d}\right) \ (the\ rotation\ related\ to\ J2)$$

$$\theta = \frac{\pi}{2} - \theta_2 \ (the\ rotation\ related\ to\ J3)$$

$$\theta_3 = -\theta \ (the\ rotation\ related\ to\ J4)$$

(Formulas credits to Ali Elouafiq)

### V-3 THE END EFFECTOR (GRIPPER):

In order to move pawns, the arm should have a gripper that enables the arm to grab things. The gripper used is fractional jaw grippe that has two jaws that enable to grab objects. These are equations that describe the torque applied by the gripper that will enable us in further steps to choose the right gripper for specific robot arm weight, and pawns mass.

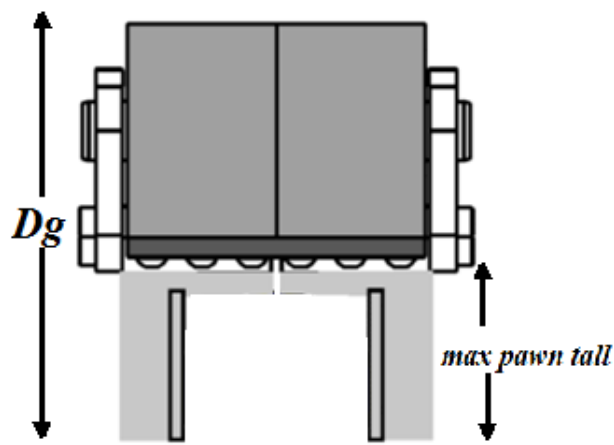

Figure V-3-1: Scheme of the gripper

(Scheme credits to [8])

Long Jaws are often required. In either case, the longer the jaw the greater the torque the gripper imposes on itself. Therefore, the torque from the grippers is;

$$\tau_{gripper} = F_{gripper} \cdot L_{jaw}$$

$$\tau_{part} = a_{total} \cdot M_{part} \cdot L_{jaw}$$

Where $L_{jaw}$ is measured from the face of the gripper to the center of gravity of the part). And $a_{total}$ is the sum of the accelerations applied on the system, and $M_{part}$ the total mass of the system.[8]

Which leads to :

$$\tau_{total} = \tau_{part} + \tau_{gripper} = a_{total} \cdot M_{part} \cdot L_{jaw} + F_{gripper} \cdot L_{jaw}$$

In the next steps of building the arm, I'm going to program a simulation that enables to test different parameters in order to choose the optimized length and mass of the gripper.

### V-4 THE ALGORITHMS:

The algorithms described in this section will enable the robot arm to do its job. The equations and formulations are not enough, since the motion should be synchronized and to avoid environment errors. Moreover, the algorithms are hardware restricted, they should be as fast as possible, without recursion, and do not use a lot of memory (32kbytes constraint).

**Start of the system: (the initialization algorithm)**

```
Declare x=0;

Declare y=0;

Declare L=/* set the length according to the current board*/;

Declare dif=/* set dif according to the current board*/;

Declare d=(15L-16dif)/32;

Declare Teta1=Teta2=Teta3=Teta=0;

Declare Xf=1/16;

Declare Yf=1/16+dif;
```

**go to x**

```
Xf=x*L/8 -L/16;

Teta1=Teta1+ Arctan(Xf/Yf);

Rotate J1 by Teta1;
```

(Algorithms credits to Ali Elouafiq)

### go to y

```
Yf= y*L/8-L/16+dif;

Teta2=Teta2+Arcsin(Yf/d);// revolves join 2

Teta=pi/2-Teta2;// revolves join 3

Teta3=-Teta; //adjusts the gripper

Rotate J2 by Teta2

Rotate J4 by Teta3

Rotate J3 by Teta

// this sequence was built in a way that the gripper doesn't
disturb pawns in the Table
```

### move from [x ,y] to [x',y']:

```
go to x

go to y

grab

Rotate J2 by -Teta2/2;

go to x'

Rotate J2 by Teta2/2;

go to y'

release
```

### move to [x ,y]:

```
go to x

go to y
```

### return to o:

```
Rotate J2 by -Teta2/2;

go to x=0

Rotate J2 by +Teta2/2;

go to y=0
```

(Algorithms credits to Ali Elouafiq)

## VI- CONCLUSION:

This study on robot arms was fulfilled to get enough material on how to design robot arm for specific needs, in the scope of the "ChessBot" project. After, comparing and benchmarking industrial different robot arms, I've came out with the right design for my project. The inverse kinematics equations were a bit tricky to solve. Moreover, the algorithms challenge was in the sequence of execution of the steps that will avoid trouble making while moving pawns. Till now, nothing implemented yet, the next step will be to test a simulation of this design using Lego Mindstorms, and trying to find the physical measures of the arm, before going to prototyping

## VIII- BIBLIOGRAPHY: